\pdfoutput=1
\documentclass[11pt]{article}

\usepackage[]{emnlp2021}

\usepackage{times}
\usepackage{latexsym}
\usepackage{graphicx}
\usepackage{tabularx}
\newcolumntype{R}[1]{>{\raggedleft\arraybackslash}p{#1}}
\usepackage{comment}
\usepackage{placeins}
\usepackage{amsmath}
\usepackage{amssymb}
\usepackage{nccmath}
\usepackage{adjustbox}
\usepackage[T1]{fontenc}
\usepackage[utf8]{inputenc}

\usepackage{microtype}

\usepackage{algorithm}
\usepackage{algpseudocode}
\usepackage{hyperref}

\usepackage{etoolbox}

\title{MANTIS at TSAR-2022 Shared Task: Improved Unsupervised Lexical Simplification with Pretrained Encoders}

\author{
    Xiaofei Li$^1$, Daniel Wiechmann$^2$, Yu Qiao$^1$, Elma Kerz$^1$ \\
    $^1$ RWTH Aachen University \\
    $^2$ University of Amsterdam \\
    \texttt{\{xiaofei.li1,yu.qiao\}@rwth-aachen.de}\\
    \texttt{d.wiechmann@uva.nl, elma.kerz@ifaar.rwth-aachen.de}
}

\AtBeginEnvironment{align}{\useshortskip}
\hypersetup{draft}
\begin{document}
\maketitle

\setlength{\abovedisplayskip}{3pt}
\setlength{\belowdisplayskip}{3pt}
\setlength{\abovedisplayshortskip}{3pt}
\setlength{\belowdisplayshortskip}{3pt}

\begin{abstract}
In this paper we present our contribution to the TSAR-2022 Shared Task on Lexical Simplification of the EMNLP 2022 Workshop on Text Simplification, Accessibility, and Readability. Our approach builds on and extends the unsupervised lexical simplification system with pretrained encoders (LSBert) system introduced in \citet{qiang2020lexical} in the following ways: For the subtask of simplification candidate selection, it utilizes a RoBERTa transformer language model and expands the size of the generated candidate list. For subsequent substitution ranking, it introduces a new feature weighting scheme and adopts a candidate filtering method based on textual entailment to maximize semantic similarity between the target word and its simplification. Our best-performing system improves LSBert by 5.9\% accuracy and achieves second place out of 33 ranked solutions.%We further explore the utility of crowdsourcing- and corpus-based measure of word prevalence for substitution ranking.

\end{abstract}

\section{Introduction}

Lexical simplification (LS) is a natural language processing (NLP) task that involves automatically reducing the lexical complexity of a given text, while retaining its original meaning \cite{shardlow2014survey,paetzold2017survey}. Since LS has a high potential for social benefit and improving social inclusion for many people, it has attracted increasing attention in the NLP community \cite{vstajner2021automatic}. LS systems are commonly framed as a pipeline of three main steps \cite{paetzold2017lexical}: (1) Complex Word Identification (CWI), (2) Substitute Generation (SG), and (3) Substitute Ranking (SR), with CWI often being treated as an independent task. 

In this paper, we present our contributions to the English track of the TSAR-2022 Shared Task on LS \cite{tsar2022}. Focusing on steps (2) and (3) in the pipeline above, the task was defined as follows: Given a sentence containing a complex word, systems should return an ordered list of “simpler” valid substitutes for the complex word in its original context. The list of simpler words (up to a maximum of 10) returned by the system should be ordered by the confidence the system has in its prediction (best predictions first). The ordered list must not contain ties. The task employed a new benchmark dataset for lexical simplification in English, Spanish, and (Brazilian) Portuguese. The gold annotations consists of all simpler substitutes suggested by crowdsourced workers and checked for quality by at least one computational linguist who is native speaker of the respective language (for details, see \citet{10.3389/frai.2022.991242}). Contributing teams were provided with a small sample with gold standard annotations as a trial dataset. For English, this trial dataset consists of 10 instances of a sentence, a target complex word and a list of substitution candidates. The English test dataset consisted of 373 instances of sentence/complex word pairs. Submission were evaluated in terms of ten performance metrics that fall into three groups: (1) MAP@K (Mean Average Precision@$K$) for $K={1,3,5,10}$ candidate words. This metric evaluates a ranked list of predicted substitutes that is matched (relevant) and not matched (irrelevant) terms against the set of the gold-standard annotations for evaluation. (2) Potential@$K$: $K={1,3,5,10}$. Potential scores quantify the percentage of instances for which at least one of the substitutions predicted is present in the set of gold annotations and (3) Accuracy@$K$@top1: $K={1,2,3}$. Accuracy scores represent the ratio of instances where at least one of the $K$ top predicted candidates matches the most frequently suggested synonym/s in the gold list of annotated candidates.

\section{System Description}

Our contributions to the TSAR shared task builds on and extends the approach to unsupervised lexical simplification with pretrained encoders – LSBert – described in \citet{qiang2020lexical} and \citet{qiang2021lsbert}. This approach leverages a pretrained transformer language models to generate context-aware simplifications for complex words. The LSBert simplification algorithm addresses two of three principal subtasks of LS: simplification candidate generation and substitution ranking. 

Our approach extends LSBert in the following ways: (1) It utilizes a RoBERTa transformer language model for simplification candidate generation and expands the size of the generated candidate list. (2) It introduces new substitution ranking methods that involve (i) a re-weighting of the ranking features used by LSBert and (ii) the adoption of equivalence scores based on textual entailment to maximize semantic similarity between the target word and its simplification. In submissions (runs) 2 and 3, we further explore the utility of crowdsourcing- and corpus-based measure of word prevalence for substitution ranking. The simplification algorithm underlying the three submissions described in this paper is shown in Algorithm 1. In the following we describe the details of simplification candidate generation (2.1), substitution ranking (2.2) and obtaining equivalence scores (2.3).

\begin{algorithm}
\caption{Lexical Simplification}
\label{algo}
% \begin{algorithmic}[1]
% \Require sentence $S$, Complex word $w$
% Replace word $w$ of S into [MASK] as $S′$
% \hspace*{\algorithmicindent} \textbf{Input} \\
\hspace*{0pt} \textbf{Input: } sentence $S$, Complex word $w$\\
\hspace*{0pt} \textbf{Output: } sorted suggestion list $word\_list$
\begin{algorithmic}[1]
\State Replace word $w$ of S into <mask> as $S'$
\State Concatenate $S$ and $S'$ using <s> and </s>
\State $p(\cdot|S,S'\setminus\{w\}) \gets RoBERTa(S,S')$
\State $scs \gets top\_probability(p(\cdot | S, S'\setminus\{w\}))$
\State $all\_ranks \gets \emptyset$
\For{each feature $f$ and its weight $c_f$}
\State $scores\gets \emptyset$
\For{each $sc\in scs$}
    \State $scores\gets scores \cup f(sc)$
\EndFor
\State $rank\gets c_f\times rank\_numbers(scores)$
\State $all\_ranks \gets all\_ranks \cup rank$
\EndFor
\State $tot\_rank \gets sum(all\_ranks)$
\State $word\_list' \gets sort\_ascending(tot\_rank)$
\State $word\_list \gets postproc(word\_list')$\\
\Return $word\_list$
\end{algorithmic}
% \end{algorithmic}
\end{algorithm}

\subsection{Simplification Candidate Generation}

During candidate generation, for each pair of sentence $S$ and complex word $w$, the LSBert algorithm first generates new sequence $S'$ in which $w$ is masked. The two sentences $S$ and $S'$ are then concatenated and fed into a pretrained transformer language  model (PTLM) to obtain the probability distribution of the vocabulary that can fill the masked position, $p(\cdot|S,S \setminus \{w\})$. The top 10 words from this distribution are considered as the list of simplification candidates.\footnote{Morphological derivations of $w$ are excluded.} Our simplification candidate generation method differs from the one used in LSBert in two ways: (1) the choice of PTLM and (2) the size of the candidate list. \citet{qiang2021lsbert} performed experiments with three BERT models: (i) BERT-based, uncased: 12-layer, 768-hidden, 12-heads, 110 M parameters. (ii) BERT-large, uncased: 24-layer, 1024-hidden, 16-heads, 340 M parameters, and (iii) BERT-large, uncased, Whole Word Masking (WWM): 24-layer, 1024-hidden, 16-heads, 340 M parameters. The results of their experiments indicated that the WWM model obtains the highest accuracy and precision. Here we extended these PTLM-experiments to include RoBERTa models \cite{liu2019roberta} and also experimented with the combined use of BERT and RoBERTa to enlarge the list of substitution candidates. The results of our experiments indicated that optimal results are obtained using the RoBERTa-md: 12-layer, 768-hidden, 12-heads, 125M parameters. To maximize the chance of obtaining at least ten suitable substitution candidates after rigorous filtering based on semantic criteria (see below), we increased the size of the candidate list generated in this step from 10 to 30 candidates.

\subsection{Substitution Ranking}
In LSBert, candidate substitutions are ranked based on four features each of which is designed to capture one aspect of the suitability of the candidate word to replace the complex word. These features are rank orders of candidate substitutions based on four scores: (1) `Pretrained LM (PTLM) prediction' ($B_{PTLM}(sc)$, in LSBert, PTLM = Bert) representing the probability derived from PTLM that the candidate substitution word $sc$ presents at the masked position given the rest of a sentence. (2) `Language model feature' ($L_{PLM}(sc)$) representing the average loss of the context of $sc$, $w_{-m}^{m}=(w_{-m}, w_{-m + 1}, \dots,w_0,\dots, w_{m - 1}, w_m)$, where $w_0=sc$. (3) `Semantic similarity' ($S(sc)$) expressed as the cosine similarity between the fastText vector of the original word and the that of the $sc$. (4) `Word frequency' ($F(sc)$) as estimated from the top 12 million texts from Wikipedia and the Children's Book Test corpus.\footnote{\url{https://github.com/google-research/bert}} In LSBert, the rank of a $sc, R(sc)$, is based on an equal weighting of these four features, as shown in equation (1) and (2). 

\begin{align}
    Score(sc) &= \frac{1}{4}\sum_{f\in\{B_{Bert}, -L_{Bert}, S, F\}} rank_f(sc)\\
    R(sc) &= rank_{Score}(sc)
\end{align}
where $rank_f:  SCS \rightarrow \mathbb{Z}$:
\[sc \mapsto |\{w\in SCS| f(w) > f(sc)\}| + 1\]
and $SCS$ is the set of all substitution candidates. 

%In LSBert, candidate substitutions are ranked based on four features each of which is designed to capture one aspect of the suitability of the candidate word to replace the complex word. These features are: (1) the `BERT prediction' (B) representing the rank derived from the probability distribution of the vocabulary corresponding to the mask word, (2) the `language model feature' (L) representing the rank based on the average loss of $W$, where $W$ denotes the context of the original word $w$, (3) the ranked `semantic similarity' (S) expressed as the cosine similarity between the fastText vector of the original word and the fastText vector of the candidate substitutions and (4) the ranked `word frequency' (F) as estimated from the top 12 million texts from Wikipedia and the Children’s Book Test corpus. In LSBert, the ranking of substitute candidates ($sc$) is based on an equal weighting of theses four features, as shown in equation 1. 

%\begin{equation}
 %   LSBert\  rank(sc) = \frac{1}{4}\sum_{i\in\{B, L, S, F\}} i(sc)
%\end{equation}

In our three submissions to the shared task, we considered three different strategies to derive the above $Score(sc)$: In the first submission (Mantis\_1), we adapted the ranking method as shown in equation (3). $c_f$ is the feature weight for feature $f$ and $c_{B_{Roberta}}=c_F=1, c_S=3$.

\begin{align}
    Score_{run 1}(sc) &= \sum_{f\in\{B_{Roberta}, S, F\}} c_f\cdot rank_f(sc)
\end{align}

This ranking method introduces a re-weighting of the features so as to (i) increase the relative importance of the semantic similarity between the target word $w$ and a substitute candidate $sc$ and (ii) decrease the relative importance of the probability-based PTLM prediction. With regard to the former, the value of $S(sc)$, corresponding to ranked cosine similarity, was increased by a factor of 3 to penalize candidates with low similarity to the target word. With regard to the latter, we decided to drop the language model feature $L_{PTLM}(sc)$ as its correlation with $B_{PTLM}(sc)$ would yield an up-weighting of the importance assigned to the probability of $sc$ to appear in the masked position. 

In the second and third submissions (Mantis\_2 and Mantis\_3), we experimented with alternative features for substitution ranking: To this end, we first computed lexical complexity scores for the sentences in the trial data for each substitution candidate using 77 indicators (see Table 2 in the appendix). All scores were obtained using an automated text analysis system developed by our group (for its recent applications, see e.g. \citet{wiechmann2022measuring} or \citet{kerz-etal-2022-pushing}). Tokenization, sentence splitting, part-of-speech tagging, lemmatization and syntactic PCFG parsing were performed using Stanford CoreNLP \citep{manning2014stanford}. We then used each feature to obtain a rank order of substitution candidates and correlated reach ranking with the rank order of substitution candidates provided in the trial data. The top-2 lexical features yielding the largest correlations with the gold standard ranking were selected for substitution ranking for Mantis\_2 and Mantis\_3, respectively. Both of these lexical features concern word prevalence (WP), i.e. they refer to the number of people who know the word: WP\textsubscript{crowd} estimates the proportion of the population that knows a given word based on a crowdsourcing study involving over 220,000 people \cite{brysbaert2019word}. WP\textsubscript{corp.SDBP} is an corpus-derived estimate of the number of books that a word appears in \cite{johns2020estimating}. The corresponding rankings were obtained as shown in equations (4) and (5): 

\begin{align}
 Score_{run 2}(sc) &= \sum_{f\in\{WP_{crowd}, Eq\}} rank_f(sc)\\
Score_{run 2}(sc) &= \sum_{f\in\{WP_{corp.SDBP}, Eq\}} rank_f(sc)
\end{align}

Apart from these WP-features, the substitution ranking in runs 2 and 3 was determined by a semantic feature, referred to as the `equivalence score' $Eq(sc)$ (see section 2.3). This score was evoked based on the consideration that semantic similarity measured by cosine similarity of embeddings is not expressive enough \cite{kim2016intent}: Any two words that are frequently used in similar contexts will have a low cosine similarity between the embeddings. Thus cosine similarity often fails to recognize antonyms, such as "fast" and "slow". The next section will provide more details on how equivalence score were obtained.  

\begin{figure*}
    \centering
    \includegraphics[width = 0.83\textwidth]{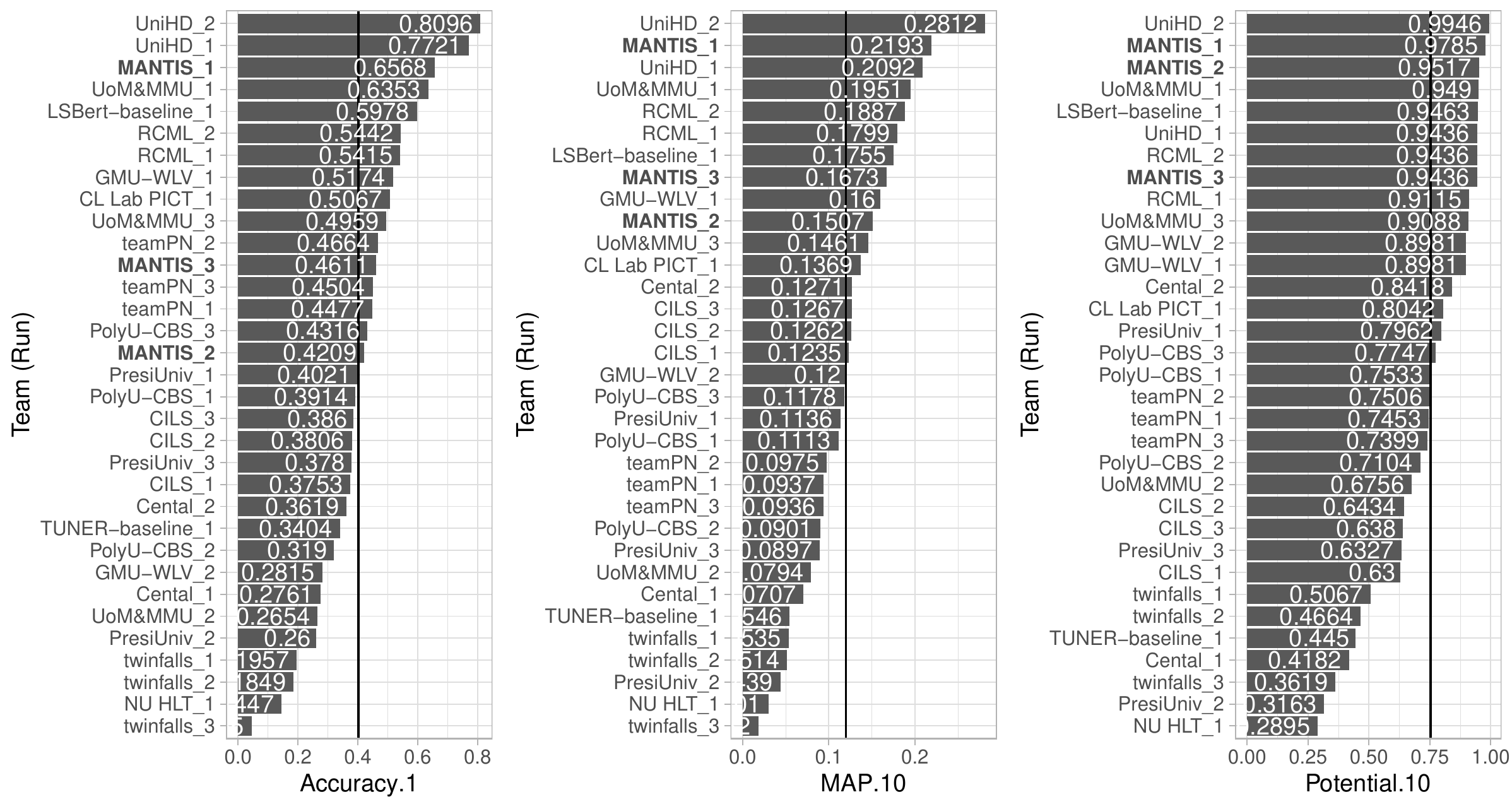}
    \caption{Performance ranking based on Accuracy, Mean Average Precision, and Potential scores (k=10). Vertical lines represent the median performance across the 33 submission for each metric. }
    \label{fig:results}
    \vspace{-5mm}
\end{figure*}

\subsection{Obtaining Equivalence Scores}

Lexical simplification needs to preserve the original meaning of the target word. As cosine similarity between embedding vectors can be too permissive, we introduced a stricter criterion based on textual entailment. To achieve this we utilized a language model explicitly trained to the natural language inference (NLI) task of evaluating logical connections between sentences. The central idea is to compute for each substitute word $sc$ a score that quantifies the textual entailment of the original sentence $S$ and its variant $S`$ that contains $sc$. Textual entailment is a directional relation between text fragment that holds whenever the truth of one text fragment follows from another text. The entailing and entailed texts are termed premise ($p$) and hypothesis ($h$), respectively. The relation between $p$ and $h$ can be one of entailment, contradictory or neutral (neither entailment nor contradictory).  To the extent that $p$ and $h$ mutually entail each other, they are considered equivalent. In this paper, the entailment scores were obtained from the `roberta-large-mnli' model from the Hugginface transformer library.\footnote{\url{https://huggingface.co/roberta-large-mnli}} Roberta-large-mnli is a RoBERTa large model fine-tuned on the Multi-Genre Natural Language Inference corpus using a masked language modeling objective \cite{N18-1101}. The entailment score is defined as the probability that $p$ entails $h$:

\begin{align}
    En(p, h) = Prob_\theta(entailment\;|\;  p, h)
\end{align}

where $\theta$ is the parameters of trained roberta-large-mnli. We quantify the degree of equivalence of two sentences (\textit{equivalence score}) as the product of the entailment scores in both directions.  For a given sentence $S$ and the corresponding simplified sentence $S'$, the equivalence score is defined as:

\begin{align}
    Eq(S, S')=En(S, S')\cdot En(S', S)
\end{align}

Apart from their use in the substitution ranking in Mantis\_2 and Mantis\_3, equivalent scores were also used in a postprocessing step in Mantis\_1: Here the list of substitution candidates was pruned after ranking by removing candidates whose equivalence scores were smaller than the mean equivalence score of all candidates. 

\section{End-to-end System Performance}

The official results across seven performance metrics\footnote{Four of the ten performance metrics, Acc@1, MAP@1, Potential@1, and Precision@1, give the same results as per their definitions.} are presented in Table \ref{tab:offResults} in the appendix (for details, see \citet{tsar2022}). As the performance metrics are strongly intercorrelated (mean correlation across all metrics = 0.920, sd = 0.071, see also Figure \ref{fig:cor} in the appendix), we focus our discussion here on the results of one metric from each of the three groups: (1) Accuracy.1, (2) MAP.10 and (3) Potential.10 (see Figure \ref{fig:results}). Our best-performing system was `Mantis\_1'. This system reached 2\textsuperscript{nd} rank on both MAP.10 and Potential.10 and 3\textsuperscript{rd} rank on accuracy. Mantis\_1 displayed an improvement over the median performance of +25.56\% on accuracy, +24.13\% on potential.10 and +9.93\% MAP.10. It outperformed the LSBert baseline by +5.9\% accuracy, +4.38 MAP.10 and 3.49\% Potential.10. The two systems whose substitution ranking was based solely on word prevalence and an equivalence score lagged behind the LSBert baseline on two of the performance metrics shown here, suggesting that the improvements of our system over LSBert was mainly due to better substitution ranking, rather than candidate selection. However, Mantis\_2 outperformed LSBert on the Potential.10 metric, suggesting that the inclusion of word prevalence can be fruitfully employed to improve LS systems. In future work, we intend to explore the role these and additional indicators of lexical sophistication for substitution ranking.  

\FloatBarrier
\bibliography{anthology,custom}

\begin{thebibliography}{17}
\expandafter\ifx\csname natexlab\endcsname\relax\def\natexlab#1{#1}\fi

\bibitem[{Brysbaert et~al.(2019)Brysbaert, Mandera, McCormick, and
  Keuleers}]{brysbaert2019word}
Marc Brysbaert, Pawe{\l} Mandera, Samantha~F McCormick, and Emmanuel Keuleers.
  2019.
\newblock Word prevalence norms for 62,000 english lemmas.
\newblock \emph{Behavior research methods}, 51(2):467--479.

\bibitem[{Davies(2008)}]{davies2008corpus}
Mark Davies. 2008.
\newblock The {C}orpus of {C}ontemporary {A}merican {E}nglish ({COCA}): 560
  million words, 1990-present.

\bibitem[{Johns et~al.(2020)Johns, Dye, and Jones}]{johns2020estimating}
Brendan~T Johns, Melody Dye, and Michael~N Jones. 2020.
\newblock Estimating the prevalence and diversity of words in written language.
\newblock \emph{Quarterly Journal of Experimental Psychology}, 73(6):841--855.

\bibitem[{Kerz et~al.(2022)Kerz, Qiao, Zanwar, and
  Wiechmann}]{kerz-etal-2022-pushing}
Elma Kerz, Yu~Qiao, Sourabh Zanwar, and Daniel Wiechmann. 2022.
\newblock Pushing on personality detection from verbal behavior: A transformer
  meets text contours of psycholinguistic features.
\newblock In \emph{Proceedings of the 12th Workshop on Computational Approaches
  to Subjectivity, Sentiment {\&} Social Media Analysis}, pages 182--194,
  Dublin, Ireland. Association for Computational Linguistics.

\bibitem[{Kim et~al.(2016)Kim, Tur, Celikyilmaz, Cao, and Wang}]{kim2016intent}
Joo-Kyung Kim, Gokhan Tur, Asli Celikyilmaz, Bin Cao, and Ye-Yi Wang. 2016.
\newblock Intent detection using semantically enriched word embeddings.
\newblock In \emph{2016 IEEE spoken language technology workshop (SLT)}, pages
  414--419. IEEE.

\bibitem[{Liu et~al.(2019)Liu, Ott, Goyal, Du, Joshi, Chen, Levy, Lewis,
  Zettlemoyer, and Stoyanov}]{liu2019roberta}
Yinhan Liu, Myle Ott, Naman Goyal, Jingfei Du, Mandar Joshi, Danqi Chen, Omer
  Levy, Mike Lewis, Luke Zettlemoyer, and Veselin Stoyanov. 2019.
\newblock Roberta: A robustly optimized bert pretraining approach.
\newblock \emph{arXiv preprint arXiv:1907.11692}.

\bibitem[{Manning et~al.(2014)Manning, Surdeanu, Bauer, Finkel, Bethard, and
  McClosky}]{manning2014stanford}
Christopher~D Manning, Mihai Surdeanu, John Bauer, Jenny~Rose Finkel, Steven
  Bethard, and David McClosky. 2014.
\newblock The stanford corenlp natural language processing toolkit.
\newblock In \emph{Proceedings of 52nd annual meeting of the association for
  computational linguistics: system demonstrations}, pages 55--60.

\bibitem[{Paetzold and Specia(2017{\natexlab{a}})}]{paetzold2017lexical}
Gustavo Paetzold and Lucia Specia. 2017{\natexlab{a}}.
\newblock Lexical simplification with neural ranking.
\newblock In \emph{Proceedings of the 15th Conference of the European Chapter
  of the Association for Computational Linguistics: Volume 2, Short Papers},
  pages 34--40.

\bibitem[{Paetzold and Specia(2017{\natexlab{b}})}]{paetzold2017survey}
Gustavo~H Paetzold and Lucia Specia. 2017{\natexlab{b}}.
\newblock A survey on lexical simplification.
\newblock \emph{Journal of Artificial Intelligence Research}, 60:549--593.

\bibitem[{Qiang et~al.(2021)Qiang, Li, Zhu, Yuan, Shi, and
  Wu}]{qiang2021lsbert}
Jipeng Qiang, Yun Li, Yi~Zhu, Yunhao Yuan, Yang Shi, and Xindong Wu. 2021.
\newblock Lsbert: Lexical simplification based on bert.
\newblock \emph{IEEE/ACM Transactions on Audio, Speech, and Language
  Processing}, 29:3064--3076.

\bibitem[{Qiang et~al.(2020)Qiang, Li, Zhu, Yuan, and Wu}]{qiang2020lexical}
Jipeng Qiang, Yun Li, Yi~Zhu, Yunhao Yuan, and Xindong Wu. 2020.
\newblock Lexical simplification with pretrained encoders.
\newblock In \emph{Proceedings of the AAAI Conference on Artificial
  Intelligence}, volume~34, pages 8649--8656.

\bibitem[{Saggion et~al.(2022)Saggion, \v{S}tajner, Ferr{\'e}s, Sheang,
  Shardlow, North, and Zampieri}]{tsar2022}
Horacio Saggion, Sanja \v{S}tajner, Daniel Ferr{\'e}s, Kim~Cheng Sheang,
  Matthew Shardlow, Kai North, and Marcos Zampieri. 2022.
\newblock Findings of the tsar-2022 shared task on multilingual lexical
  simplification.
\newblock In \emph{Proceedings of TSAR workshop held in conjunction with EMNLP
  2022}.

\bibitem[{Shardlow(2014)}]{shardlow2014survey}
Matthew Shardlow. 2014.
\newblock A survey of automated text simplification.
\newblock \emph{International Journal of Advanced Computer Science and
  Applications}, 4(1):58--70.

\bibitem[{{\v{S}}tajner(2021)}]{vstajner2021automatic}
Sanja {\v{S}}tajner. 2021.
\newblock Automatic text simplification for social good: Progress and
  challenges.
\newblock \emph{Findings of the Association for Computational Linguistics:
  ACL-IJCNLP 2021}, pages 2637--2652.

\bibitem[{\v{S}tajner et~al.(2022)\v{S}tajner, Ferr\'{e}s, Shardlow, North,
  Zampieri, and Saggion}]{10.3389/frai.2022.991242}
Sanja \v{S}tajner, Daniel Ferr\'{e}s, Matthew Shardlow, Kai North, Marcos
  Zampieri, and Horacio Saggion. 2022.
\newblock \href {https://doi.org/10.3389/frai.2022.991242} {{Lexical
  simplification benchmarks for English, Portuguese, and Spanish}}.
\newblock \emph{Frontiers in Artificial Intelligence}, 5.

\bibitem[{Wiechmann et~al.(2022)Wiechmann, Qiao, Kerz, and
  Mattern}]{wiechmann2022measuring}
Daniel Wiechmann, Yu~Qiao, Elma Kerz, and Justus Mattern. 2022.
\newblock Measuring the impact of (psycho-) linguistic and readability features
  and their spill over effects on the prediction of eye movement patterns.
\newblock \emph{arXiv preprint arXiv:2203.08085}.

\bibitem[{Williams et~al.(2018)Williams, Nangia, and Bowman}]{N18-1101}
Adina Williams, Nikita Nangia, and Samuel Bowman. 2018.
\newblock A broad-coverage challenge corpus for sentence understanding through
  inference.
\newblock In \emph{Proceedings of the 2018 Conference of the North American
  Chapter of the Association for Computational Linguistics: Human Language
  Technologies, Volume 1 (Long Papers)}, pages 1112--1122. Association for
  Computational Linguistics.

\end{thebibliography}
\bibliographystyle{acl_natbib}

\newpage
\onecolumn
\appendix
% \FloatBarrier

% \onecolumn
\section{Appendix}
\label{sec:appendix}

% \begin{comment}
    
% \begin{figure*}
%     \centering
%     \includegraphics[width = 1.0\textwidth]{fig/SharedTask.png}
%     \caption{
%     NOTES\\1. Candidates were not filtered out when ranking by linguistic features, because this would lead to worse performance on the trial dataset. \\ 2. if the correlation(feature, Morphological.MeanLengthWord) > 0, then rank in increasing order else in decreasing order.\\
%     The model “roberta-large-mnli” is actually a classification model. It classifies sentences pairs as CONTRADICTION,  NEUTRAL or ENTAILMENT. The entailment score is actually the probability that a sentence pair is classified as ENTAILMENT. 
% }
%     \label{fig:my_label}
% \end{figure*}
% \end{comment}

\begin{table}[ht!]
\setlength{\tabcolsep}{2pt}
\caption{Official results across 7 performance metrics (Acc@1, MAP@1, Potential@1, and Precision@1 give the same results as per their definitions)}
\adjustbox{max width=\textwidth}{
\begin{tabular}{|c|lc||ccc||cc||cc|}
\hline
Rank & Team            & Run & ACC@1  & ACC@1.1 & ACC@3.1 & MAP@3  & MAP@10 & Pot@3 & Pot@10 \\
\hline
1    & UniHD           & 2   & 0.8096 & 0.4289     & 0.6863     & 0.5834 & 0.2812 & 0.9624      & 0.9946       \\
2    & UniHD           & 1   & 0.7721 & 0.4262     & 0.571      & 0.509  & 0.2092 & 0.89        & 0.9436       \\
3    & \textbf{MANTIS}          & 1   & \textbf{0.6568} & 0.319      & 0.5388     & 0.473  & 0.2193 & 0.8766      & 0.9785       \\
4    & UoM\&MMU        & 1   & 0.6353 & 0.2895     & 0.5308     & 0.4244 & 0.1951 & 0.8739      & 0.949        \\
5    & \textbf{LSBert-baseline} & 1   & \textbf{0.5978} & 0.3029     & 0.5308     & 0.4079 & 0.1755 & 0.823       & 0.9463       \\
6    & RCML            & 2   & 0.5442 & 0.2359     & 0.4664     & 0.3823 & 0.1887 & 0.831       & 0.9436       \\
7    & RCML            & 1   & 0.5415 & 0.2466     & 0.4691     & 0.3716 & 0.1799 & 0.8016      & 0.9115       \\
8    & GMU-WLV         & 1   & 0.5174 & 0.2493     & 0.4477     & 0.3522 & 0.16   & 0.7533      & 0.8981       \\
9    & CLLabPICT       & 1   & 0.5067 & 0.2064     & 0.4021     & 0.3278 & 0.1369 & 0.7265      & 0.8042       \\
10   & UoM\&MMU        & 3   & 0.4959 & 0.2439     & 0.4235     & 0.3273 & 0.1461 & 0.756       & 0.9088       \\
11   & teamPN          & 2   & 0.4664 & 0.1823     & 0.3378     & 0.2743 & 0.0975 & 0.6729      & 0.7506       \\
12   & \textbf{MANTIS}          & 3   & 0.4611 & 0.2117     & 0.4235     & 0.3227 & 0.1673 & 0.7747      & 0.9436       \\
13   & teamPN          & 3   & 0.4504 & 0.1769     & 0.3297     & 0.2676 & 0.0936 & 0.6648      & 0.7399       \\
14   & teamPN          & 1   & 0.4477 & 0.1769     & 0.3297     & 0.2666 & 0.0937 & 0.6621      & 0.7453       \\
15   & PolyU-CBS       & 3   & 0.4316 & 0.2064     & 0.3297     & 0.2683 & 0.1178 & 0.6139      & 0.7747       \\
16   & \textbf{MANTIS}          & 2   & 0.4209 & 0.1662     & 0.3565     & 0.2745 & 0.1507 & 0.7131      & 0.9517       \\
17   & PresiUniv       & 1   & 0.4021 & 0.1581     & 0.3002     & 0.2603 & 0.1136 & 0.6568      & 0.7962       \\
18   & PolyU-CBS       & 1   & 0.3914 & 0.1823     & 0.3002     & 0.2576 & 0.1113 & 0.5924      & 0.7533       \\
19   & CILS            & 3   & 0.386  & 0.1957     & 0.3083     & 0.2603 & 0.1267 & 0.5656      & 0.638        \\
20   & CILS            & 2   & 0.3806 & 0.1903     & 0.3083     & 0.2597 & 0.1262 & 0.563       & 0.6434       \\
21   & PresiUniv       & 3   & 0.378  & 0.1474     & 0.2573     & 0.2277 & 0.0897 & 0.5656      & 0.6327       \\
22   & CILS            & 1   & 0.3753 & 0.201      & 0.3109     & 0.2555 & 0.1235 & 0.5361      & 0.63         \\
23   & Cental          & 2   & 0.3619 & 0.1152     & 0.2788     & 0.2573 & 0.1271 & 0.6541      & 0.8418       \\
24   & TUNER-baseline  & 1   & 0.3404 & 0.142      & 0.1823     & 0.1706 & 0.0546 & 0.4343      & 0.445        \\
25   & PolyU-CBS       & 2   & 0.319  & 0.1447     & 0.2573     & 0.1973 & 0.0901 & 0.512       & 0.7104       \\
26   & GMU-WLV         & 2   & 0.2815 & 0.0804     & 0.2493     & 0.1899 & 0.12   & 0.563       & 0.8981       \\
27   & Cental          & 1   & 0.2761 & 0.1313     & 0.2117     & 0.1635 & 0.0707 & 0.378       & 0.4182       \\
28   & UoM\&MMU        & 2   & 0.2654 & 0.1367     & 0.268      & 0.182  & 0.0794 & 0.4906      & 0.6756       \\
29   & PresiUniv       & 2   & 0.26   & 0.1018     & 0.1554     & 0.135  & 0.0439 & 0.3136      & 0.3163       \\
30   & twinfalls       & 1   & 0.1957 & 0.0509     & 0.1233     & 0.1175 & 0.0535 & 0.3485      & 0.5067       \\
31   & twinfalls       & 2   & 0.1849 & 0.0643     & 0.1367     & 0.1182 & 0.0514 & 0.3565      & 0.4664       \\
32   & NUHLT           & 1   & 0.1447 & 0.067      & 0.1179     & 0.0902 & 0.0301 & 0.26        & 0.2895       \\
33   & twinfalls       & 3   & 0.0455 & 0.0107     & 0.0455     & 0.037  & 0.0182 & 0.1474      & 0.3619\\   
\hline
\end{tabular}
}
\label{tab:offResults}
\end{table}
% \twocolumns
\begin{table}[]
    \centering
       \caption{An example instance from the trial dataset with gold annotation candidate list provided by the organizers}
    \def\arraystretch{1.5} 
    \small
\begin{tabular}{|p{2cm}|p{5cm}|}
\hline
         Sentence     & A Spanish government source, however, later said that banks able to cover by themselves losses on their toxic property assets will not be forced to remove them from their books while it will be compulsory for those receiving public help.  \\
         \hline
         Complex word & compulsory \\
         \hline
        Gold annotations & mandatory, mandatory, mandatory, mandatory, mandatory, mandatory, mandatory, mandatory, mandatory, mandatory, mandatory, required, required, required, required, required, required, required, essential, forced, important, manadatory, necessary, obligatory, unavoidable\\
         \hline
    \end{tabular}
 
    \label{tab:my_label}
\end{table}

\begin{table}
  \centering
  \setlength{\tabcolsep}{2pt}
  \caption{Overview of the 77 features considered for Substitution Ranking}
  \label{tab:features}
    \begin{tabular}{|l|c|l|}
		\hline
		Feature group & N & Examples/description \\
		\hline
		Lexical Sophistication & 14    &  Mean length/word, \\
		Density and Diversity &&N Words on NGSL,\\
  %&&NDW&Num. diff words (NDW)\\
  %&&cNDW&Corrected number of different words\\
  
%		&&TTR&Type-Token Ratio (TTR)\\
		&&Corrected TTR\\
%		&&rTTR&Root TTR\\
 %           &&log TTR&Logarithmic TTR\\
		\hline
		Register-based  &   25  & N-gram freq.\\
	N-gram Frequency	&          & (N = 1-5)\\
		&         &  five subcorpora \\
		&         & from COCA\\
  &         & \cite{davies2008corpus}\\
\hline
Psycholinguistic & 38&Age of Acquisition, \\
 & &Word Prevalence \\
  & &(corpus-based), \\
   & &Word Prevalence  \\
  & &(crowdsourced) \\
 \hline
     \end{tabular}
\end{table}

\begin{figure}
    \centering
    \includegraphics[width = 1\textwidth]{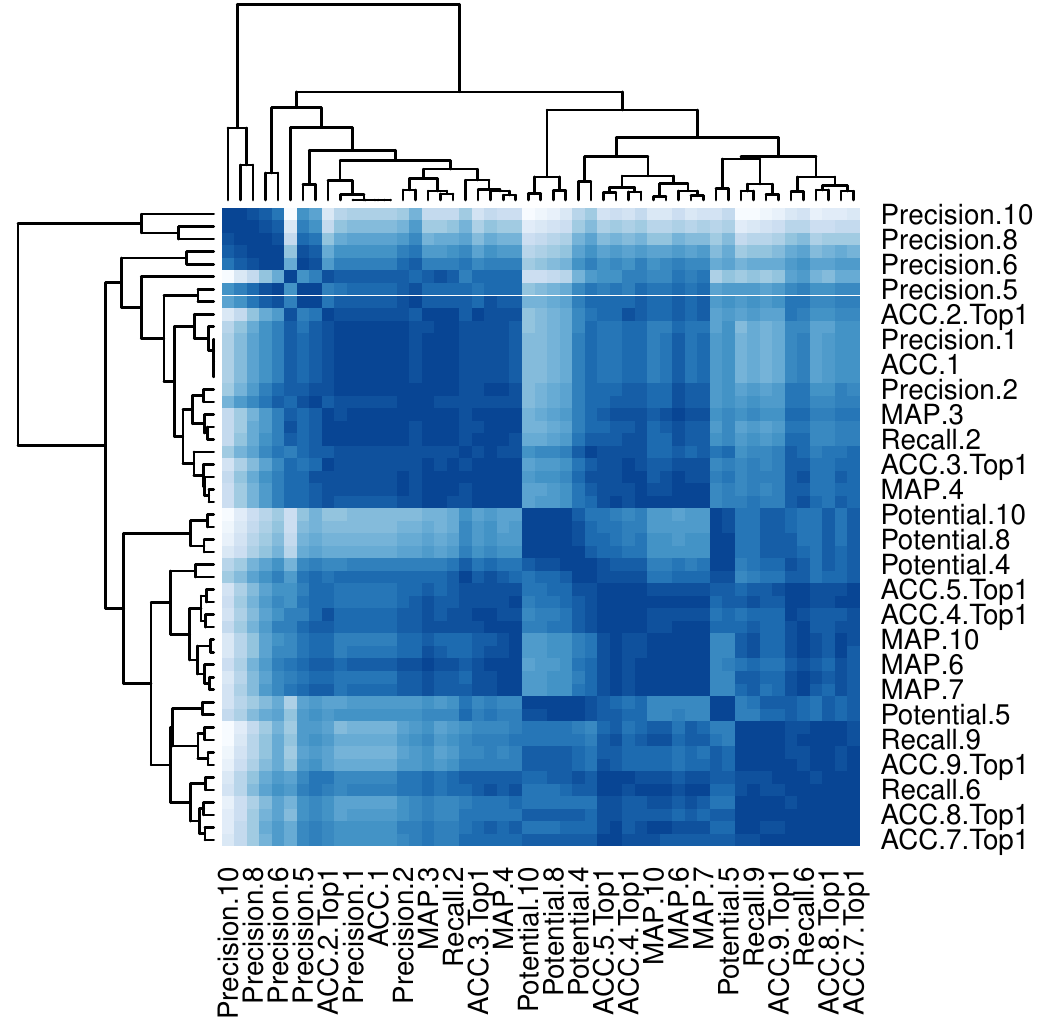}
    \caption{Heatplot of intercorrelations (Pearson r) of evaluation metrics (ns=50). Mean r = 0.920, sd = 0.071. Ranks of the three runs submitted were constant across metrics (run1 = 3, run2 = 12, run3 = 16).}
    \label{fig:cor}
\end{figure}

\end{document}